\pdfoutput=1

\documentclass[11pt]{article}

\usepackage[]{acl}

\usepackage{times}
\usepackage{latexsym}
\usepackage{graphicx} 
\usepackage{booktabs}
\usepackage{multirow}
\usepackage{siunitx}
\usepackage{subcaption}
\usepackage{tikz}
\usepackage{amsmath}
\usetikzlibrary{positioning,chains,fit,shapes,calc}

\usepackage{newfloat}

\usepackage[T1]{fontenc}

\usepackage[utf8]{inputenc}

\usepackage{microtype}
\newcommand{\pattern}[1]{\textit{#1}}

\usepackage{bibentry}

\title{Breaking BERT: Evaluating and Optimizing Sparsified Attention}
\usepackage{algorithm}
\usepackage{algorithmic}

\newcommand{\ceil}[1]{\left\lceil #1 \right\rceil}

\definecolor{n_blue}{RGB}{76, 114, 176}
\definecolor{n_red}{RGB}{196, 78, 82}
\definecolor{n_green}{RGB}{85, 168, 104}

\tikzset{cross/.style={cross out, draw=black, minimum size=2*(#1-\pgflinewidth), inner sep=0pt, outer sep=0pt},
cross/.default={4pt}}

\tikzset{crosssm/.style={cross out, draw=black, minimum size=2*(#1-\pgflinewidth), inner sep=0pt, outer sep=0pt},
crosssm/.default={2.5pt}}

\newcommand{\circlegr}{\tikz\draw[n_green,fill=n_green]  (0,0) circle (.4ex); }
\newcommand{\circlered}{\tikz\draw[n_red,fill=n_red]  (0,0) circle (.4ex); }
\newcommand{\crossblue}{\tikz\draw  (0,0) node[cross,rotate=45, n_blue, very thick] {}; }
\newcommand{\squareblue}{\tikz\draw[n_blue,fill=n_blue]  (0,0) rectangle (0.12,0.12); }
\newcommand{\crossbluesm}{\tikz\draw  (0,0) node[crosssm,rotate=45, n_blue, very thick] {}; }

\usepackage{newfloat}
\usepackage{listings}
\floatstyle{ruled}
\newfloat{listing}{tb}{lst}{}
\floatname{listing}{Listing}
 \author{\quad Siddhartha Brahma\thanks{~~Equal contribution.} \quad Polina Zablotskaia\footnotemark[1] \\  {\bf David Mimno}\\
 Google Research\\
 \texttt{\{sidbrahma,polinaz,mimno\}@google.com}
}

\begin{document}

\definecolor{myblue}{RGB}{80,80,160}
\definecolor{mygreen}{RGB}{80,160,80}

\maketitle

\begin{abstract}
Transformers allow attention between all pairs of tokens, but there is reason to believe that most of these connections---and their quadratic time and memory---may not be necessary.
But which ones?
We evaluate the impact of sparsification patterns with a series of ablation experiments.
First, we compare masks based on syntax, lexical similarity, and token position to random connections, and measure which patterns reduce performance the least.
We find that on three common finetuning tasks even using attention that is at least 78\% sparse can have little effect on performance if applied at later transformer layers, but that applying sparsity throughout the network reduces performance significantly. 
Second, we vary the degree of sparsity for three patterns supported by previous work, and find that connections to neighbouring tokens are the most significant.
Finally, we treat sparsity as an optimizable parameter, and present an algorithm to learn  degrees of neighboring connections that gives a  fine-grained control over the accuracy-sparsity trade-off while approaching the performance of existing methods.
\end{abstract}

\section{Introduction}

BERT~\cite{bert} models achieve high performance at the cost of increased computation and decreased interpretability. The combination of multiple learned self-attention heads has the capacity to carry a significant amount of information between all pairs of tokens, but it appears that many of these potential connections are either unused or unnecessary \cite{lu2021influence}. For example, many attention heads empirically correspond to relatively simple patterns such as ``attend to the following token'' or ``attend to the first token'' \cite{clark2019what}, and many can be removed without affecting performance significantly \cite{michel}.
One state-of-the-art model for long sequences~\cite{zaheer} uses a specific combination of three such patterns: globally connected tokens, neighbour connections, and random connections.

The success of these sparsity patterns raises several questions.
While probing studies have provided insight into the behavior of individual heads, the aggregate effect of attention is less clear, even in models that induce sparsity~\cite{meister2021sparse}.
Which patterns are most important?
Are there other patterns that could perform equally well or better?
And do specific parameterizations of patterns affect performance? Is it possible to learn these sparse patterns while optimizing for sparsity and accuracy?
We address these questions with three experiments.

First, rather than trying to explain what attention \textit{is} doing, we attempt to rule out what attention is \textit{not} doing by applying aggressive sparsity patterns while finetuning from a standard pre-trained model.
We apply a series of fixed masks with fixed sparsity to each input text based on an instance-specific graph structure, including parse trees, semantic similarity networks, linear-chain neighbours, and random graphs.
Our goal in such drastic model alterations is \textit{not} to improve performance but to measure which interventions damage performance the least:
if a model cannot ``rewire'' an existing model to accommodate a sparser network of interactions, the missing interactions must have been necessary in some way.
We find that most patterns are similar, but linear-chain neighbour connections are the most robust over three commonly used finetuning tasks.

In addition, we find that sparse attention masks can be applied to the final layers of a BERT model with little impact on performance, indicating that these layers can be sparsified or even eliminated. In contrast, applying the same masks to earlier layers has a more significant effect on performance, indicating that dense connections are more important at the early stages of encoding.

Second, we investigate the effect of varying the degree of sparsity of individual patterns when they are used in combination, by exploring the space of combinations of global, random, and neighbour connections used in BigBird \cite{zaheer}.
Again, we find that eliminating neighbour connections has the most drastic effect among these patterns.

Finally, having established the importance of neighbour connections, we present a method inspired by LDPC codes from coding theory that learns the degree of connectivity in the neighbour connections while optimizing for sparsity. We show that our method gives a more fine-grained control over the trade-off between accuracy and sparsity than methods using fixed sparsity patterns like BigBird. It also achieves higher accuracies than BigBird in the high sparsity regime. 
To summarize, our contributions are:
\begin{itemize}
\item We investigate the attention mechanism in BERT by applying aggressive sparsity patterns. The \pattern{Neighbour} and \pattern{Random} patterns are least damaging to the model, indicating that syntactic or semantic meaning is not the only information the model learns.
\item We further explore the best patterns with an ablation study centered on the BigBird model. We try various combinations of sparsity patterns and node degrees, again confirming the importance of the \pattern{Neighbour} pattern.
\item We propose a method to learn the degrees of sparse \pattern{Neighbour} patterns that allows for more fine-grained control over models with desired accuracy and sparsity requirements. Our results provide guidance to developers of transformer-based models to anticipate the effect of any desired degree of sparsity.
\end{itemize} 

\section{Related Work}

There has been considerable work on sparsifying attention.
Attention heads are known to ``specialize'' \cite{clark2019what}, and removing heads that were initially available can have surprisingly little effect relative to using fewer heads from the start~\cite{voita-etal-2019-analyzing, michel}.
Attention can also be reformulated as conditional expectations filtered through sparse networks of tokens \cite{ren2021combiner}, resulting in reductions in time and memory.
Other approaches include the Star Transformer~\cite{Guo}, Sparse Transformer~\cite{child},  Log-Sparse Transformer~\cite{li2019enhancing} and the Block Sparse Transformer~\cite{qiu2019blockwise}. All these methods use specific predefined sparsity patterns to mask the full attention matrix.

In searching for effective sparse patterns, we perform ablation experiments in which we remove all connections except those that correspond to a specific pattern.
We compare networks based on syntactic structure~\cite{tenney, hewitt,Coenen,kermit, zhang-etal-2016-top} as well as semantic similarity~\cite{jawahar-etal-2019-bert, kelly} and positional neighbours~\cite{clark2019what,zaheer}.

There is a strong connection between graphs and attention. Graph neural networks (GNNs) have become increasingly popular, and have been used as an alternative to transformers for language modeling~\cite{bastings-etal-2017-graph, marcheggiani-titov-2017-encoding}. It has been noted that attention weights in transformer models can be thought of as a ``soft'' graphical structure where all token nodes are fully connected, but the edge weights vary based on learned patterns. In both the transformer and GNN context, performance and scalability depend critically on finding a simple attention matrix, equivalent to applying a sparse graph between tokens. But because there are combinatorially many graph topologies, searching over all possible graphs to find the optimal sparsity structure would be prohibitively difficult. 

In contrast to predefined  sparsity patterns, a second line of work tries to derive adaptive sparsity patterns based on data. Adaptive Sparse Transformer~\cite{correia2019adaptively} uses Entmax instead of Softmax; Routing Transformer~\cite{roy2021efficient} uses non-negative matrix factorization on the attention matrix. Another recent paper, Reformer~\cite{kitaev2020reformer}, optimizes attention by using locally sensitive hashing (LSH) to only compute a few dominant dot products rather than materializing the whole attention matrix. These approaches usually learn sparser structures more adapted to the characteristics of a particular task. They also allow for a higher level of sparsity control, responsive to a learning objective. 
Using an approach like this can help us better understand how the model learns, especially in combination with a static pattern. 
Our third experiment on learning patterns while optimizing sparsity is most directly comparable to Adaptive Attention Span~\cite{sukhbaatar2019adaptive}, but we learn global sparsity patterns specific to the \textit{task} rather than being specific to each \textit{instance}.


\section{Transformer sparsity and LDPC codes}

Our objective is to find well performing sparse connectivity patterns between tokens.
The quadratic scaling of attention produces blowup in time and memory that can be burdensome even for short sequences.
We know that much of this work is not needed, but which parts?
The best pattern would be simple, requiring little to no information about the specific text sequence; as sparse as possible; but maintaining as much of the important information flow as possible.

\begin{figure}
\begin{center}
    \includegraphics[scale=0.16]{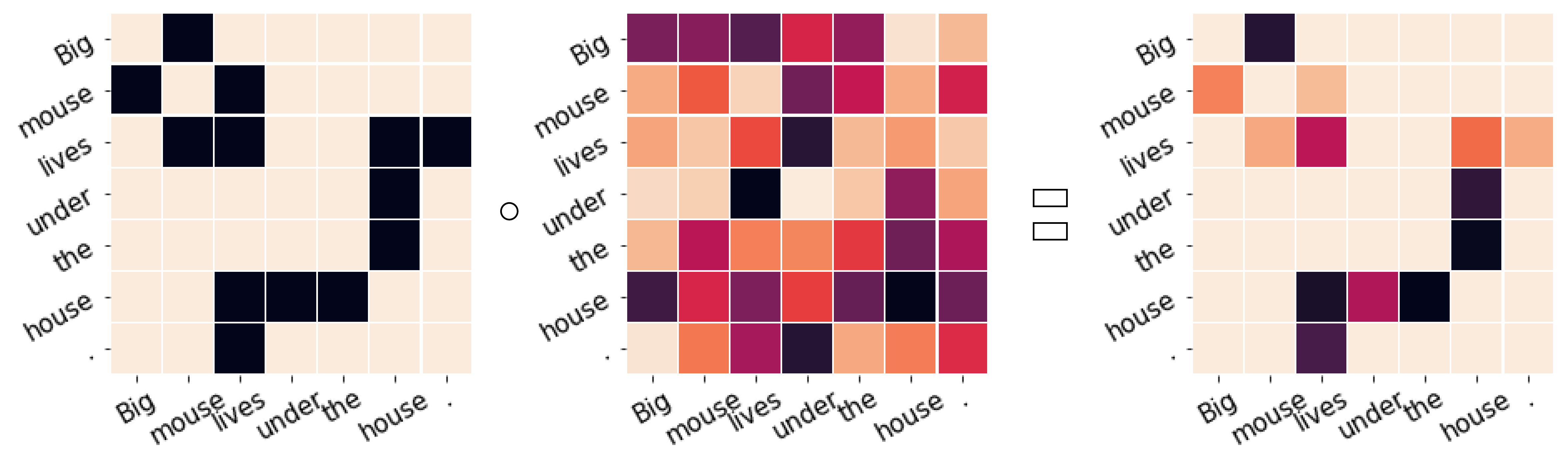}
    \caption{\footnotesize{Hadamard product between a sparsity pattern $M$ and the full attention matrix $QK^T$.}}
    \label{fig:hadamard}
\end{center}
\end{figure}

As a Transformer-based~\cite{vaswani2017attention} model, BERT uses the self-attention mechanism~\cite{attention_ex1, attention_ex2, attention_ex3} as a key component to transform a sequence of representations $X = [x_1,x_2,...,x_n]$ of length $n$. If $d$ is the dimension of each of the representations, the following function computes a new set of representations
\begin{equation}
Attention(Q,K,V)  = \sigma\left(\frac{QK^T}{\sqrt{d}}\right)V 
\end{equation}
Here each of the matrices $Q,K,V$ are linear transformations (each of dimension $n\times d$) of the matrix of representations $X$, respectively called the query, key and value matrices. The time complexity of computing the attention matrix $QK^T$ is $O(n^2 d)$. 
If the attention matrix is sparsified by the Hadamard product between a binary sparsification mask $M$ (of dimension $n\times n$ specifying the sparse pattern) and $QK^T$~(Fig.~\ref{fig:hadamard}), then the complexity of the attention mechanism reduces to $O((1-\varsigma(M))n^2 d)$, where $\varsigma(M)$ is the sparsity of $M$ which is the fraction of entries in $M$ equal to 0.  

One can think of $QK^T$ as a matrix encoding information about the interaction of all the tokens in a sequence as defined by  the $n^2$ pairwise dot-products. The goal of discovering sparse attention masks $M$ then is to find more compact encodings of this interaction using fewer pairwise terms. But the search space for $M$ is exponentially large; since each position is a binary variable it is $O(2^{n^2})$. To make the problem more tractable, we take inspiration from coding theory and in particular from LDPC (low density parity check) codes \cite{codingtheorybook}.
For LDPC codes, the task is to discover sparse $M$ that can be used as parity check matrices for encoding data. Using theoretical results that show that given a specific degree distribution, random matrices have mostly equivalent performance, the search then reduces to finding sparse degree distributions and then sampling random matrices from an optimal degree distribution.


We apply the same reasoning to search for good  sparse $M$ (random or otherwise) to act as  ``encoders'' of the attention information in transformers. Note that, in contrast to LDPC codes where each of the $n$ positions is equivalent, positional information in a  sequence of tokens of natural language (and hence their transformer representations) is important. This observation leads to a series of research questions:
\begin{itemize}
    \item Are sparse random graphs optimal for language, which exhibits sequential and hierarchical structure? We address this issue by comparing a random graph to a selection of structured sparsity patterns, based on syntax, lexical similarity, and positional proximity.
    \item Can we measure difference between specific sparsity patterns, and are they sensitive to degree? We select three patterns used in the BigBird model, enumerate combinations of node-degrees and analyze their performance.
    \item Within the best set of patterns, is the optimal pattern learnable? Similar to LDPC codes, we formulate a problem of learning degree distributions on these patterns and fine-tune models with an objective that optimizes for sparsity (which is a function of the degree distribution) as well as accuracy.
\end{itemize}

\section{Data and Sparsity patterns}
Coding theory implies that random graphs have good performance for coding, but does this result apply to connections between tokens in language data?
Because we know that language has structure, linguistically aware graphs may be more effective than simple or random connections.
We evaluate four increasingly complicated sparse graph structures for three tasks from the GLUE benchmark \cite{glue} of increasing levels of complexity. 
\begin{table}[]
\caption{\footnotesize{Average properties by dataset: cosine similarity between pairs of tokens in a sentence, number of tokens in a sentence(s), total number of unique tokens in a dataset, examples that we used for finetuning (dropping those for which we could not extract syntax trees).}}\label{data_prop:1}
\resizebox{8cm}{!}{
\begin{tabular}{*7l}
\toprule
Metric &  \multicolumn{2}{c}{CoLA} & \multicolumn{2}{c}{SST-2} & \multicolumn{2}{c}{MNLI} \\
\midrule
{}   & Train   & Val.   & Train  & Val.   & Train   & Val. \\
Similarity($*10^{-4}$)    &61  &69  &79  &10 & 83      & 80   \\
$N$ tokens         & 9. &9.44 &10.12  &20.92 &34.93    & 34.26  \\
Vocab size               & 5415 & 1863 & 13814  &4287 &70582    & 13294 \\
$N$ examples           &8551 &1043 &67349 &872 &392702   &9815\\
\bottomrule
\end{tabular}
}
\end{table}
%
SST-2~\cite{glue} is a sentiment prediction task that is easier than the other two tasks as measured by predictive accuracy. Bag-of-words baselines are relatively strong for this task compared to the others.
CoLA~\cite{glue}  measures a system's ability to predict whether a sentence is grammatical. As this task depends on linguistic features such as word order, it cannot be solved through purely lexical methods.
MNLI~\cite{glue} records the logical relationship between two sentences as implies, contradicts, or neutral. Validation accuracy is lower than the other tasks and, like CoLA, it cannot be solved using lexical methods alone. 

\begin{figure*}
    \centering
    \includegraphics[scale=.38]{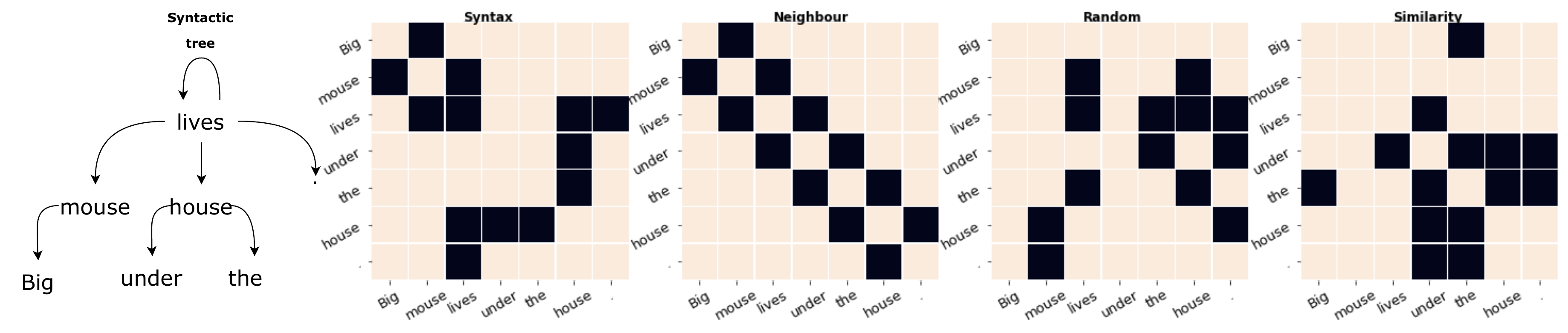}
    \caption{\footnotesize{\textbf{Sparsity graphs}. We sparsify attention matrices with 4 different graphs represented as sparse adjacency matrices $M$. \textit{Syntax} matrices come directly from the syntactic trees that we infer for each sentence in the datasets. \textit{Neighbour} matrix is formed by keeping attentions from each token only to its immediate neighbours. \textit{Random} is a matrix consisting of randomly sampled token attentions, but the sparsity $(S)$ is the same as in the \pattern{syntax} attention. Finally, the \textit{Similarity} matrix has connections where the cosine similarity scores were high enough to reach the same overall sparsity level as the \pattern{Syntax} graph.} }
    \label{fig:graphs_diagram}
\end{figure*}

Given an instance of input text, we generate four instance-specific graphs to represent different sparsity patterns~(Fig.~\ref{fig:graphs_diagram}) as described below. To make the task more difficult, we apply the same sparsity pattern to the attention matrices of all the layers of the transformer except the last one. 


\paragraph{Syntax tree.} In this pattern we use a syntax tree produced by a graph-based dependency parser trained on Universal Dependencies \cite{dependencies} that uses a small transformer -- distilled from multilingual BERT \cite{bert} -- as the neural feature extractor. We then convert the trees to the corresponding symmetrical adjacency matrices with a density of $\frac{2}{n}$ or equivalently, a sparsity of $S = 1- \frac{2}{n}$, where $n$ is the number of tokens in a sentence. For inputs with multiple sentences (e.g. MNLI), we compute the syntax trees for both sentences and then combine them. We have to also account for the word-piece tokenization that is expected in a transformer model. We resolve it via computing the offsets between the original parents of the token and the new parents after tokenization, and if a parent token has been split into multiple word-pieces, we assume the parent to be just the first piece. Further, for every piece of a child token that has been split we assume the same parent. We define the remaining sparsity patterns to match the sparsity $S$ of the syntax tree.

\paragraph{Word similarity.} In this pattern we use pretrained GloVe embeddings~\cite{pennington-etal-2014-glove} to evaluate the cosine similarity between words in an input text. We rank word pairs by their cosine similarity and greedily select edges until we reach the level of sparsity $S$ to match the more constrained syntax graph. We make sure that the edges are symmetrical and since the syntax tree has an odd number of edges (due to the root pointing to itself), we add one extra connection. Just like in the syntactic trees we also adjust the graph to the word-piece tokens. Although this method is simpler than producing parse trees, it still requires extra computation and memory to store embeddings and to compute input-specific graphs.

\paragraph{Neighbour.} This graph represents a linear chain topology in which each note is connected to the previous node and the following node. It is attractive as it could be ``hard coded'' without any variation or dependency on the content of the input. The sparsity of the \pattern{Neighbour} graph is $1- \frac{2}{n} + \frac{2}{n^2} \approx S$. The intuition for this pattern comes from~\cite{Roy, zaheer} where it is shown to be effective.

\paragraph{Random.} Inspired by LDPC codes we include a graph with randomly chosen edges, to match the sparsity $S$ of the syntax tree graph.
This is the only graph that is not symmetric.
Unlike the \pattern{neighbour} graph, each token has two random neighbours, so while this sparsity pattern is computationally simple it is more complex than the linear chain.







In the datasets we consider, mean sparsity increases with the average sequence length, from $S \approx 78\%$ for CoLA to $S\approx 94\%$ for MNLI.

\section{Comparison of sparsity patterns}\label{res}
\begin{figure}
\begin{center}
    \includegraphics[scale=0.3]{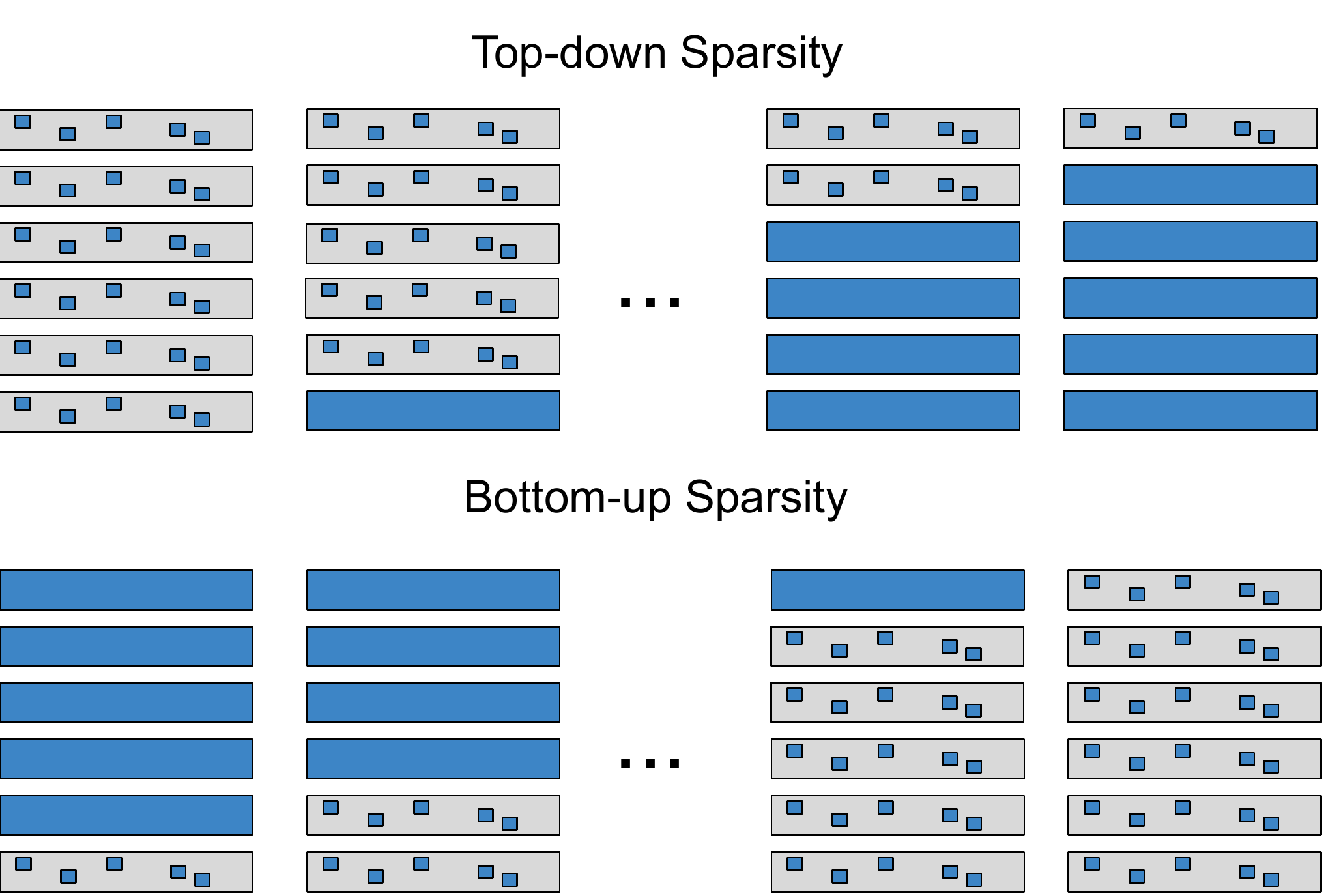}
    \caption{\footnotesize{In the top-down setting, for each layer we replace dense attention (filled box) with sparse attention (small squares) from the final layer to that layer. In the bottom-up setting we do the opposite.}}
    \label{fig:top-bottom}
\end{center}
\vspace{-3pt}
\end{figure} 
\begin{table*}[h]
\caption{\textbf{Graph properties} provide insight as to why certain graphs show better accuracy. \textit{Diameter} is the longest shortest path between two tokens in a sentence. \textit{ Shortest path} is an average across all shortest paths in a sentence. \textit{Number of components} is only relevant to the unconnected graphs. We average over all the examples in the train/validation sets.}\label{data_prop:1}
\resizebox{16cm}{!}{
\begin{tabular}{c|c|c|c|c|c|c|c|c|c}
\toprule
Metric &  \multicolumn{3}{c|}{CoLA} & \multicolumn{3}{c|}{SST-2}& \multicolumn{3}{c}{MNLI}\\
\midrule
{}       & Diameter   & Shortest path & Num components  & Diameter   & Shortest path & Num components & Diameter   & Shortest path & Num components  \\
Random        &$\infty$ &$\infty$ & 3.75 / 3.82 &$\infty$ &$\infty$ & 4.12 / 6.47 & $\infty$ &$\infty$ & 9.46 / 9.31\\
Syntax       & 3.81 / 3.93 &2.19 / 2.23 &1 &3.76 / 5.74 &2.13 / 2.89  &1 & 5.19 / 5.10 &2.70 / 2.67 & 1\\
Similarity       & $\infty$ &$\infty$ & 2.58 / 2.71 &$\infty$ &$\infty$ & 3.59 / 7.24 &$\infty$ &$\infty$ &11.68 / 11.38\\
Neighbour & 10.32 / 10.67 & 4.11 / 4.22 &1 & 12.32/24.16 & 4.77/8.72 &1 &  18.90/18.53 & 6.97/6.84 &1 \\
\bottomrule
\end{tabular}
}
\end{table*}

\begin{figure*}
    \centering
    \includegraphics[scale=0.37]{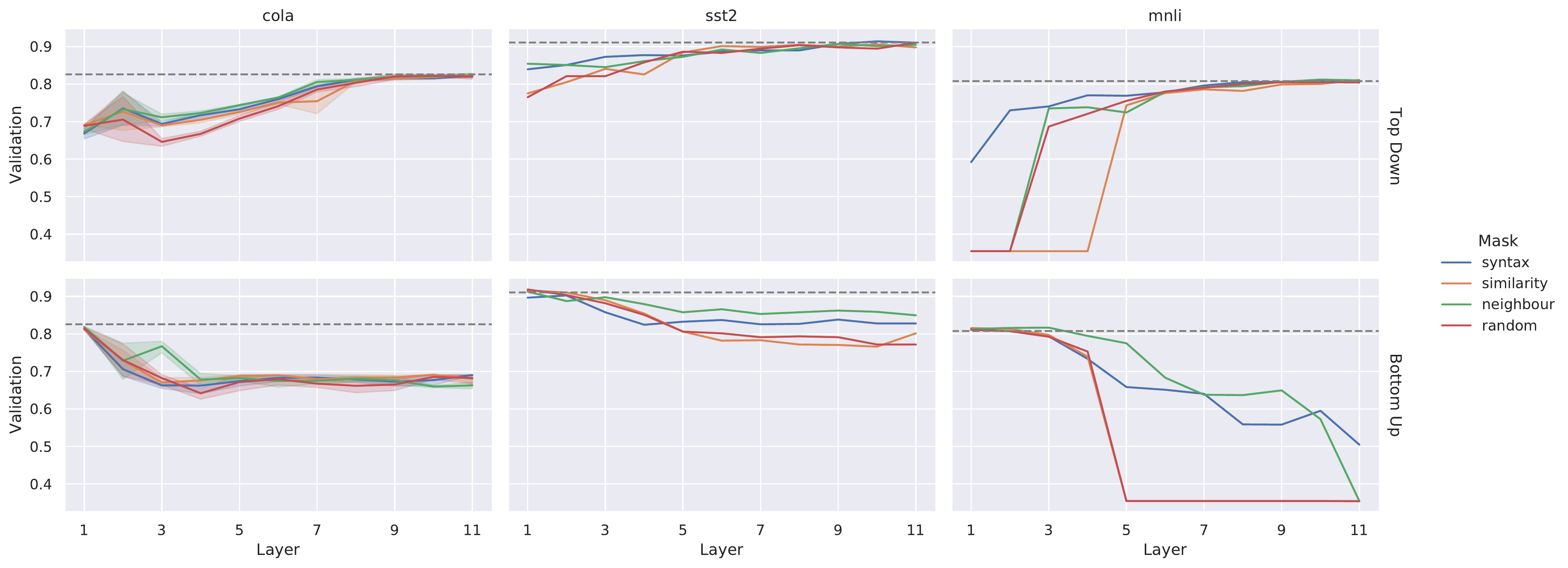}
    \caption{We apply four different mask topologies to multiple layers of BERT models. In the top row we apply the mask to each layer at or above the specified layer, in the bottom row we apply the mask to each layer at or below the specified layer.  For the latter, \pattern{Similarity} is almost indistinguishable from \pattern{Random} for MNLI, and performance drops to a baseline of majority class prediction quickly. Dashed lines show values for full attention.}
    \label{fig:facet_results}
\end{figure*}

In separate experiments we enforce attention sparsity in two directions, starting from the bottom layer and moving towards the top, and starting from top layer and moving towards the bottom.
For the \emph{Top Down} experiments, layers starting from layer $l$ are sparsified, while for the \emph{Bottom Up} experiments, the layers up to  layer $l$ are sparsified; where $l \in [1,2,\cdots, 12]$. We initialize from a pretrained BERT-Base checkpoint \cite{bert}, apply the sparsity patterns to the transformer layers and fine-tune it for each of the tasks, using standard hyperprameter settings~(see Appendix \ref{finetune_hparam}). We report validation accuracy for each of the tasks.
We do this because there is evidence that the layers of a transformer are not interchangeable, but rather serve specific roles.
For example \cite{jawahar-etal-2019-bert} observe that bottom and middle layers are associated with surface and syntactic features, while top layers encode semantic features.
If it is true that layers have different effects on the hidden representations, then we should see different results from these two patterns. 

For the smaller CoLA dataset we record five independent runs for each setting.
For the Top Down experiments, for up to four top layers only one pair of mask topologies was distinguishable from dense attention at $p<0.05$, for the others any difference was consistent with simple random variation.
As we go further down, \pattern{Random} becomes significantly worse ($p < 0.01$); at the very lowest levels runs are so variable that we cannot establish statistical significance.
For the Bottom Up pattern, using sparse attention up to various levels results in no statistically significant difference except for \pattern{Neighbour}, especially at the third level.
As we have 264 testable configurations, we do not include multiple runs with the same parameters for the two larger tasks. Performance declines more quickly when we enforce sparsity for lower layers (Bottom Up) than when we sparsify the same number of top layers (Top Down). Full, dense attention may be more critical at the early stages of encoding, but the Neighbor pattern is most able to maintain performance. 

In general, there was surprisingly little differentiation between graph topologies. \pattern{Neighbour} is most similar to \pattern{Syntax}. \pattern{Similarity} is closer to \pattern{Random}.
When there is differentiation, \pattern{Random} is most often worse.
However, it is important to note that we are constructing \pattern{Random} graphs that have the same level of sparsity as the \pattern{Syntax} graph.
While we focus on the most comparable case for the purposes of this paper, early results suggested that a more dense \pattern{Random} graph may have better performance while still maintaining a high level of sparsity.
Sparsity has less effect for SST-2 than CoLA and MNLI.
It is possible that the information that needs to propagate between tokens to get high performance on SST-2 is simple enough that six layers are sufficient, so that even a small number of random connections make little difference in performance relative to dense attention.
MNLI training becomes unstable when sparsity is applied to more than a few layers, dropping to a majority-class baseline.

\paragraph{Relation with graph properties}
To better understand why certain patterns achieve better performance we show average graph properties in Table~\ref{data_prop:1}.
MNLI has higher values than the other two datasets for all three metrics, possibly due to longer sentences and more complex syntactic structures. The ratio between \pattern{Syntax} and \pattern{Neighbour} shortest paths is also higher in MNLI, which might explain why \pattern{Syntax} is comparable to the \pattern{Neighbour} graph only in MNLI.

\begin{figure*}[h]
\centering
\begin{subfigure}[b]{0.63\textwidth}
 \centering
   \includegraphics[width=\textwidth]{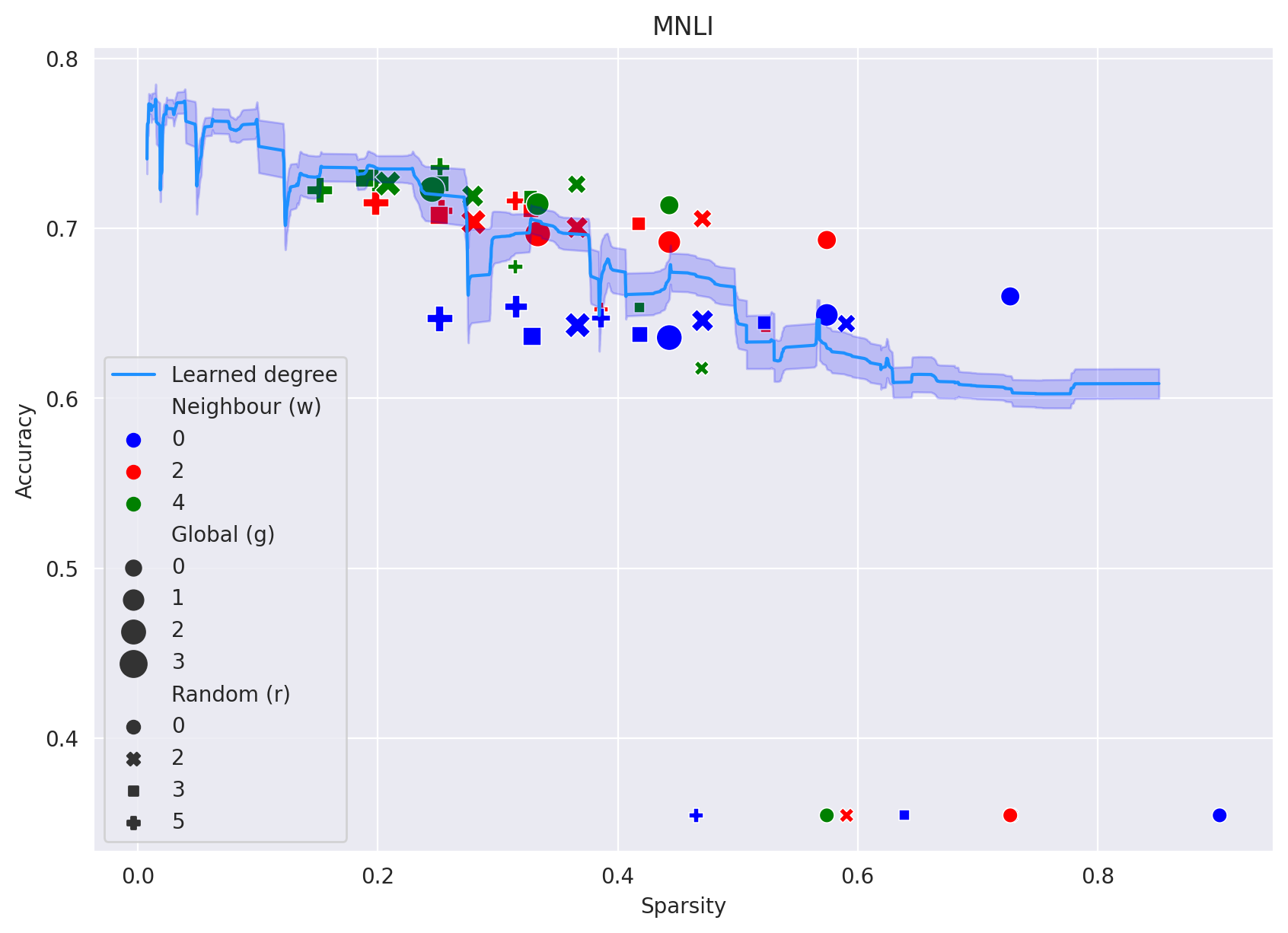}
\end{subfigure}
\hfill
\begin{minipage}[b]{0.33\textwidth}
  \begin{subfigure}[b]{\linewidth}
    \centering
     \includegraphics[width=\textwidth]{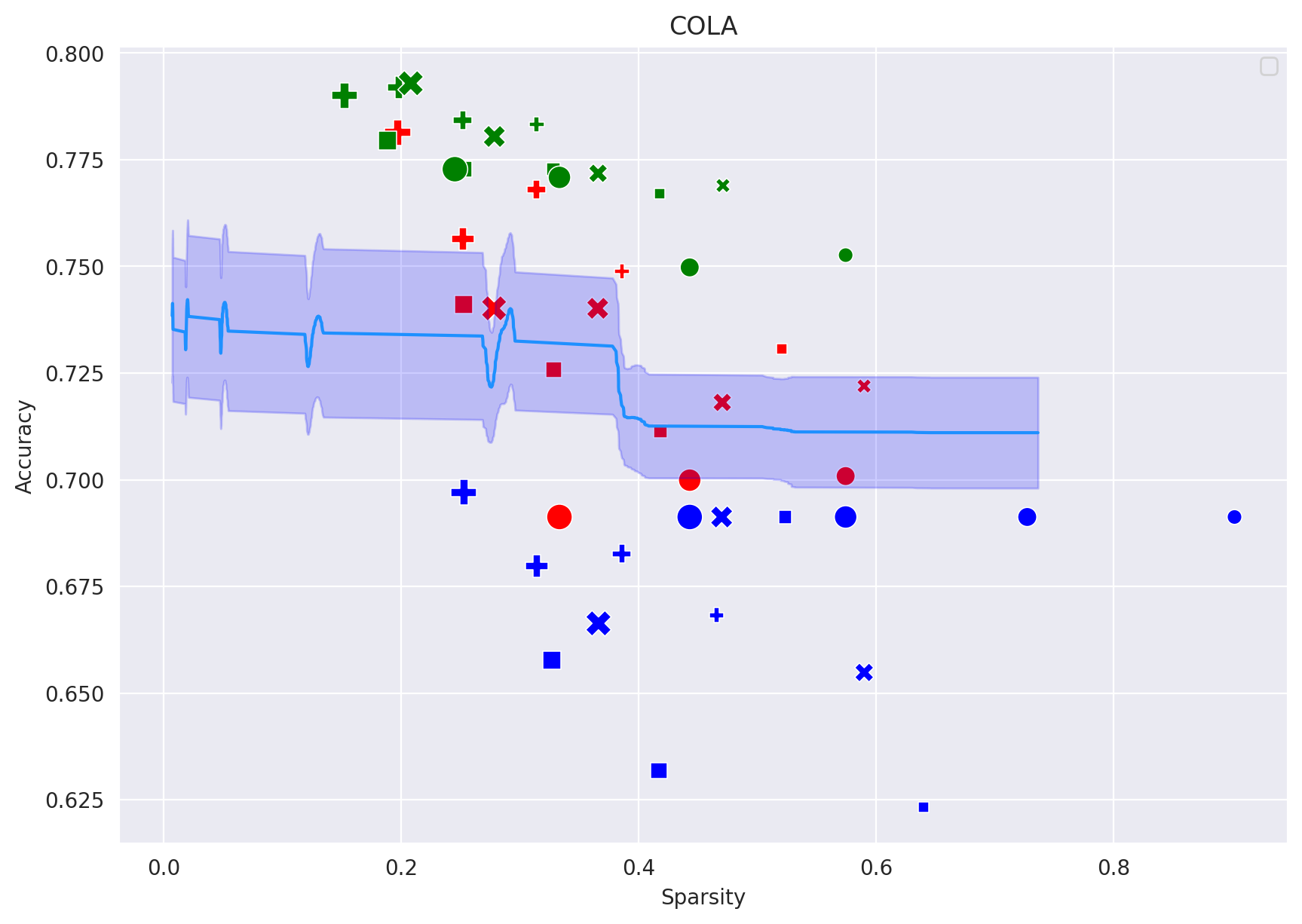}
  \end{subfigure}\\[\baselineskip]
  \begin{subfigure}[b]{\linewidth}
    \centering
    \includegraphics[width=\textwidth]{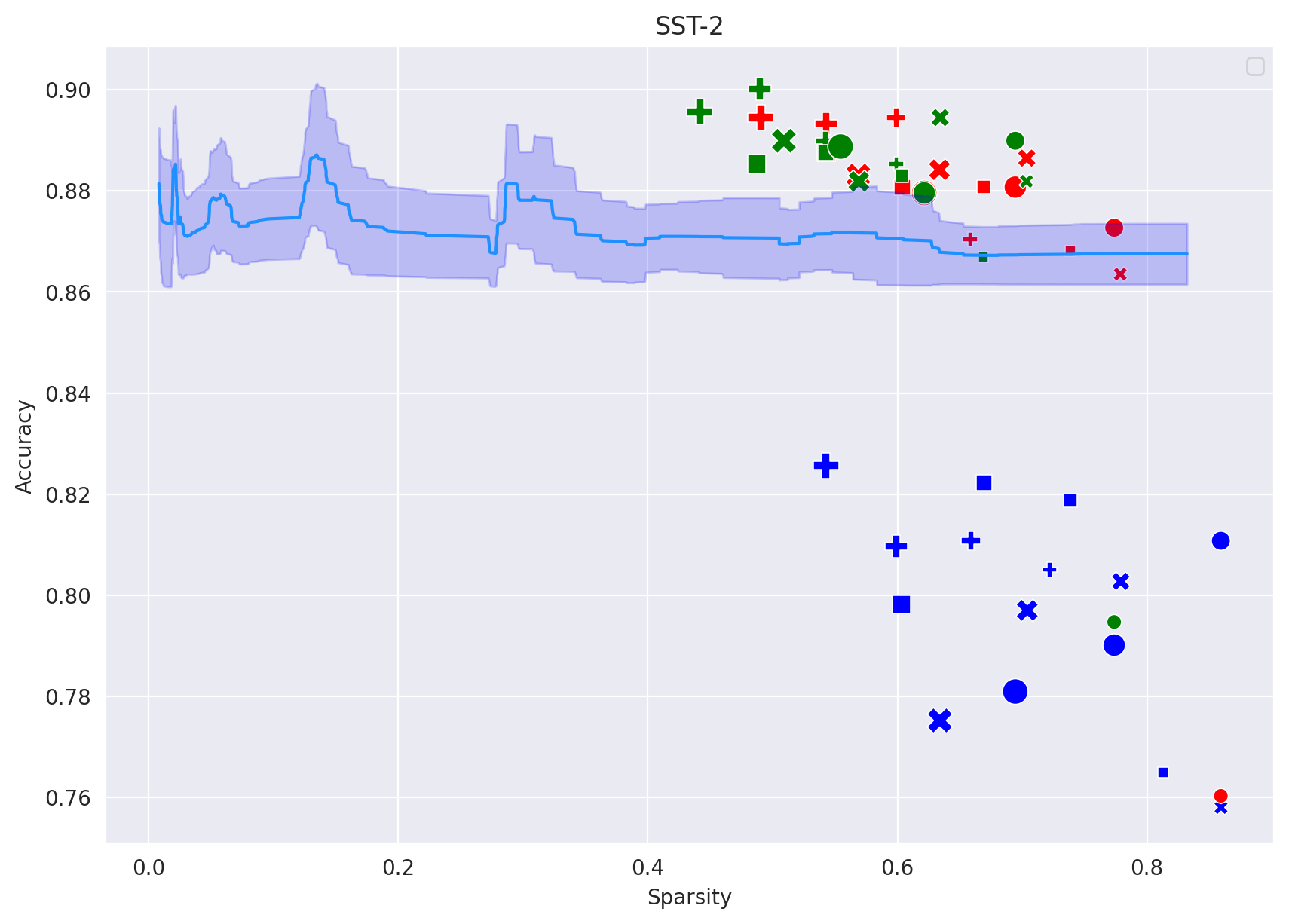}
  \end{subfigure}
\end{minipage}
\caption{\textbf{Learned sparsity vs. fixed-sparsity configurations \protect\footnotemark} Points represent fixed-sparsity configurations. For learned sparsity we estimate continuous mean and confidence regions for sparsity and validation accuracy from several training trajectories with a rolling window of size 10. Results for fixed-sparsity are reported from a single finetuned checkpoint.}\label{fig:learn_sp}
\end{figure*}

\section{Varying the degree of fixed-sparsity patterns}
The previous section compared patterns that vary in the type of information used (syntax, lexical similarity, position, random), but keep the degree of sparsity constant to make comparisons fair.
In this section we choose the simplest and highest performing of these (neighbour and random) and evaluate combinations of these with varying  degrees of sparsity.
We note that these patterns are also used in the BigBird model, which is a simple and yet effective approach to Transformer sparsification.
This model includes at its core a \pattern{Neighbour} pattern. BigBird consists of a set of $g$ global tokens attending on all parts of the sequence; all tokens attending to a window of $w$ local neighboring tokens; all tokens attending to a set of $r$ random tokens. For the GLUE experiments BigBird authors empirically chose $g = 2, w = 3, r = 3$; we are not aware of any investigation into the relative importance of the three patterns or how performance depends on the associated degree hyperparameters. As in the previous experiments, we initialize with a densely pre-trained checkpoint and apply a fixed sparsification at the finetuning stage.
This is a more difficult setting than the experiments in \citet{zaheer}, where they use sparse attention during both pre-training and finetuning, as our goal is to analyze sparsity patterns rather than to achieve state-of-the-art results.
For each configuration we calculate sparsity and accuracy.
Sparsity varies by configuration and by dataset due to differences in average sentence length.
Results for each tested configuration are shown as points in Figure~\ref{fig:learn_sp}.

\paragraph{Neighbour connections are the most important} Increasing the window size $w$ of neighbour connections improves model performance. Even configurations with only neighbour-based graphs have higher accuracy than any graph without neighbour connections. As shown on the Figure~\ref{fig:learn_sp} when $w = \{2, 4\}, g = 0, r = 0$ ( \circlegr, \circlered)  we still observe that  model performance is competitive or better than the model with configuration $w = 0, g = 3, r = 5$ ( \crossblue ) for example. 

\paragraph{Global tokens are the second most important pattern} When we set $g = 0, w = 0$  we observe for MNLI and CoLA that the models with only random connections are close to random performance (\squareblue, \crossbluesm). For MNLI, the absence of global token connections leads to random performance even with the neighbour pattern. Natural language inference inputs are longer, and are designed to require longer-range dependencies. With more tokens the potential diameter of a graph could be larger, but connections to global tokens reduce the maximum minimum path between two tokens to $2$. 

\section{Learning Sparse Degree Distributions}
\label{lsdd}

In previous sections we found that neighbour connections are powerful and that increasing the degree of node connections in sparse graphs significantly increases performance. 
In this section, inspired by how degree distributions are optimized for LDPC codes, we consider whether we can learn the degrees of node connections by optimizing for accuracy \textit{and} sparsity.


Since for all six experiments we observed the \pattern{Neighbour} pattern to be associated with the least drop in the accuracy, we will optimize the degrees of nodes in the \pattern{Neighbour} pattern. As mentioned in~\cite{zaheer} a great deal of information about a token can be derived from its neighboring tokens. From a linguistic perspective, attending to neighboring tokens is syntactically relevant and from a graph perspective it increases graph connectivity. Unlike existing work on adaptive sparsity (\cite{sukhbaatar2019adaptive}), our sparse graphs do not depend on details of an instance and can be computed without reference to data.

\footnotetext{For SST-2, the point associated with $g=0, r=0, w=0$ has accuracy 50\% and is omitted to improve the visual perspective.}


This problem is difficult because it involves discrete optimization with an exponentially large solution space.
To learn degrees that are length agnostic, we learn the fractional degrees $\delta_h^{\ell}$ for head $h$ in layer $\ell$, such that the corresponding degree is $\ceil{\delta_h^{\ell} n}$, where $n$ is the number of tokens in a sentence (or the maximum degree). Thus the attention matrix in head $h$ in layer $\ell$ is masked by each token attending to its $\ceil{\delta_h^{\ell} n}$ closest neighbours. If the original finetuning task optimizes for the model parameters $\Theta$ by minimizing $Loss(\Theta)$, we instead minimize  
\begin{equation}
    Loss'(\Theta, \{\delta_h^\ell\}) + \lambda \frac{1}{HL}\sum_{\ell=1}^{L}\sum_{h=1}^H \delta_h^{\ell}
    \label{eq:degopt}
\end{equation}
where $H$ and $L$ are the number of heads and layers in the transformer. The second term is a regularization term representing the average density of the attention matrices over all the layers, $\lambda$ being the sparsity loss coefficient. In this way, we optimize both for accuracy and sparsity, learning the  degrees of the sparsity pattern in the process. The $\delta_h^{\ell}$ are parameterized as sigmoid() of free variables. 



As before we run all the experiment with the same three tasks from the GLUE benchmark: SST-2, CoLA and MNLI. As with previous experiments, we do not sparsify the last layer in order to keep the connections with the prediction CLS token. We show the results computed on the validation dataset for our model and BigBird in Figure~\ref{fig:learn_sp}. While training our models, we record the validation accuracy and sparsity after each epoch. 
We aggregate the sequences of accuracy-sparsity pairs from 20 different experiments -- 4 different sparsity loss coefficients $\lambda =\{0.5, 1.0, 2.0, 5.0\}$ and 5 different initial sparsity levels $\{0.27, 0.37, 0.5, 0.62, 0.73\}$. As a final step of output preprocessing we have cleaned the collected points by removing the accuracy outliers and in Figure~\ref{fig:learn_sp} we demonstrate the mean and standard deviation computed across a window of size 10. We refer to our method as \textit{Learned Sparsity}.

\paragraph{Learned Sparsity allows for continuous sparsity control} As shown in  Fig.~\ref{fig:learn_sp} learned sparsity gives more finegrained control over the accuracy-sparsity trade-off curve than fixed sparsity patterns. Users can pick checkpoints with desired levels of accuracy-sparsity more effectively. 
\paragraph{Learned Sparsity performs better at high sparsity levels} Learned sparsity achieves notably better accuracy than fixed-sparsity settings when sparsity is greater than 0.8. This is especially true for SST-2, where Learned Sparsity models achieve accuracy >0.86 while fixed-sparsity accuracy drops below 0.82.
\paragraph{Some fixed-sparsity settings outperform learned sparsity} Although all fixed-sparsity configurations should be reachable under the learned objective, they sometimes show more accuracy, e.g. when only the neighbour pattern is used ($r=0$ and $g=0$) for SST-2. This discrepancy may be due to the difficulty of optimizing the sum of two competing objectives in \eqref{eq:degopt}. Also, although the $\delta_h^{\ell}$ are smooth parameters, mapping them to integer degrees may destabilize optimization. 
But when $g>0$, it is also natural to expect fixed-sparsity performance to be better since global tokens connections can have a big impact on longer sentences.

\paragraph{Learned degrees are similar across heads and layers} The degree distributions learned by Learned Sparsity are close to uniform. For example, for one model for MNLI that achieves an accuracy of 0.68, the fractional degrees have a mean of 0.56 and s.d. of 0.002. This may be related to the way the fractional degrees are parameterized.

\paragraph{Pre-training with learned sparsity is promising} While we focus in this work on finetuning dense models, we also find that pre-training with learned sparsity shows promising results. In the appendix (Fig.~\ref{fig:pretraining}) we provide MNLI results for the accuracy and degree distributions after pre-training with the learned sparsity framework.

\section{Conclusion}

Inspired by LDPC codes, we evaluate sparse attention graph topologies that most reduce density while preserving transformer performance.
We first evaluate whether random connections are as effective as more information-rich patterns at the same rate of sparsity.
We find that syntactic relationships and lexical similarity have little effect, but proximity relationships (connecting neighbouring tokens) lose the least information.
Second, we vary node degree for a combination of sparsity patterns and find that, again, neighbour graphs have the single largest effect on performance.
There does not appear to be one single optimal architecture, but rather the best sparsity pattern appears to depend on the task: simpler, more lexical tasks like SST-2 may be fine with a simpler neighbour topology, while tasks requiring more complicated joint inference over multiple sections may require more global connections.
Finally, we obtain a more fine-grained perspective on the trade-off between sparsity and performance by learning the distribution of degrees for attention heads finetuning.
While optimizing for both sparsity and performance is difficult, these results imply a lower bound and that sparser, higher performing topologies should exist.
Although Transformer architectures have revolutionized NLP and many other fields, they are inherently wasteful and opaque.
These experiments point towards new possibilities in designing better, faster, and cheaper models.




\bibliography{anthology,custom}

\begin{thebibliography}{34}
\expandafter\ifx\csname natexlab\endcsname\relax\def\natexlab#1{#1}\fi

\bibitem[{Bastings et~al.(2017)Bastings, Titov, Aziz, Marcheggiani, and
  Sima{'}an}]{bastings-etal-2017-graph}
Jasmijn Bastings, Ivan Titov, Wilker Aziz, Diego Marcheggiani, and Khalil
  Sima{'}an. 2017.
\newblock \href {https://doi.org/10.18653/v1/D17-1209} {Graph convolutional
  encoders for syntax-aware neural machine translation}.
\newblock In \emph{Proceedings of the 2017 Conference on Empirical Methods in
  Natural Language Processing}, pages 1957--1967, Copenhagen, Denmark.
  Association for Computational Linguistics.

\bibitem[{Cheng et~al.(2016)Cheng, Dong, and Lapata}]{attention_ex1}
Jianpeng Cheng, Li~Dong, and Mirella Lapata. 2016.
\newblock \href {http://arxiv.org/abs/1601.06733} {Long short-term
  memory-networks for machine reading}.
\newblock \emph{CoRR}, abs/1601.06733.

\bibitem[{Child et~al.(2019)Child, Gray, Radford, and Sutskever}]{child}
Rewon Child, Scott Gray, Alec Radford, and Ilya Sutskever. 2019.
\newblock \href {http://arxiv.org/abs/1904.10509} {Generating long sequences
  with sparse transformers}.
\newblock \emph{CoRR}, abs/1904.10509.

\bibitem[{Clark et~al.(2019)Clark, Khandelwal, Levy, and
  Manning}]{clark2019what}
Kevin Clark, Urvashi Khandelwal, Omer Levy, and Christopher~D. Manning. 2019.
\newblock What does bert look at? an analysis of bert's attention.
\newblock In \emph{BlackBoxNLP@ACL}.

\bibitem[{Coenen et~al.(2019)Coenen, Reif, Yuan, Kim, Pearce, Vi{\'{e}}gas, and
  Wattenberg}]{Coenen}
Andy Coenen, Emily Reif, Ann Yuan, Been Kim, Adam Pearce, Fernanda~B.
  Vi{\'{e}}gas, and Martin Wattenberg. 2019.
\newblock \href {http://arxiv.org/abs/1906.02715} {Visualizing and measuring
  the geometry of {BERT}}.
\newblock \emph{CoRR}, abs/1906.02715.

\bibitem[{Correia et~al.(2019)Correia, Niculae, and
  Martins}]{correia2019adaptively}
Gon{\c{c}}alo~M Correia, Vlad Niculae, and Andr{\'e}~FT Martins. 2019.
\newblock Adaptively sparse transformers.
\newblock \emph{arXiv preprint arXiv:1909.00015}.

\bibitem[{Devlin et~al.(2019)Devlin, Chang, Lee, and Toutanova}]{bert}
Jacob Devlin, Ming-Wei Chang, Kenton Lee, and Kristina Toutanova. 2019.
\newblock \href {https://doi.org/10.18653/v1/N19-1423} {{BERT}: Pre-training of
  deep bidirectional transformers for language understanding}.
\newblock In \emph{Proceedings of the 2019 Conference of the North {A}merican
  Chapter of the Association for Computational Linguistics: Human Language
  Technologies, Volume 1 (Long and Short Papers)}, pages 4171--4186,
  Minneapolis, Minnesota. Association for Computational Linguistics.

\bibitem[{Guo et~al.(2019)Guo, Qiu, Liu, Shao, Xue, and Zhang}]{Guo}
Qipeng Guo, Xipeng Qiu, Pengfei Liu, Yunfan Shao, Xiangyang Xue, and Zheng
  Zhang. 2019.
\newblock \href {http://arxiv.org/abs/1902.09113} {Star-transformer}.
\newblock \emph{CoRR}, abs/1902.09113.

\bibitem[{Hewitt and Manning(2019)}]{hewitt}
John Hewitt and Christopher~D. Manning. 2019.
\newblock \href {https://doi.org/10.18653/v1/N19-1419} {{A} structural probe
  for finding syntax in word representations}.
\newblock In \emph{Proceedings of the 2019 Conference of the North {A}merican
  Chapter of the Association for Computational Linguistics: Human Language
  Technologies, Volume 1 (Long and Short Papers)}, pages 4129--4138,
  Minneapolis, Minnesota. Association for Computational Linguistics.

\bibitem[{Jawahar et~al.(2019)Jawahar, Sagot, and
  Seddah}]{jawahar-etal-2019-bert}
Ganesh Jawahar, Beno{\^\i}t Sagot, and Djam{\'e} Seddah. 2019.
\newblock \href {https://doi.org/10.18653/v1/P19-1356} {What does {BERT} learn
  about the structure of language?}
\newblock In \emph{Proceedings of the 57th Annual Meeting of the Association
  for Computational Linguistics}, pages 3651--3657, Florence, Italy.
  Association for Computational Linguistics.

\bibitem[{Kelly et~al.(2020)Kelly, Xu, Calvillo, and Reitter}]{kelly}
M.~Kelly, Yang Xu, Jes{\'u}s Calvillo, and D.~Reitter. 2020.
\newblock Which sentence embeddings and which layers encode syntactic
  structure?
\newblock In \emph{CogSci}.

\bibitem[{Kitaev et~al.(2020)Kitaev, Kaiser, and Levskaya}]{kitaev2020reformer}
Nikita Kitaev, {\L}ukasz Kaiser, and Anselm Levskaya. 2020.
\newblock Reformer: The efficient transformer.
\newblock \emph{arXiv preprint arXiv:2001.04451}.

\bibitem[{Li et~al.(2019)Li, Jin, Xuan, Zhou, Chen, Wang, and
  Yan}]{li2019enhancing}
Shiyang Li, Xiaoyong Jin, Yao Xuan, Xiyou Zhou, Wenhu Chen, Yu-Xiang Wang, and
  Xifeng Yan. 2019.
\newblock Enhancing the locality and breaking the memory bottleneck of
  transformer on time series forecasting.
\newblock \emph{Advances in Neural Information Processing Systems},
  32:5243--5253.

\bibitem[{Lu et~al.(2021)Lu, Wang, Mardziel, and Datta}]{lu2021influence}
Kaiji Lu, Zifan Wang, Piotr Mardziel, and Anupam Datta. 2021.
\newblock Influence patterns for explaining information flow in bert.
\newblock In \emph{NeurIPS}.

\bibitem[{Marcheggiani and Titov(2017)}]{marcheggiani-titov-2017-encoding}
Diego Marcheggiani and Ivan Titov. 2017.
\newblock \href {https://doi.org/10.18653/v1/D17-1159} {Encoding sentences with
  graph convolutional networks for semantic role labeling}.
\newblock In \emph{Proceedings of the 2017 Conference on Empirical Methods in
  Natural Language Processing}, pages 1506--1515, Copenhagen, Denmark.
  Association for Computational Linguistics.

\bibitem[{Meister et~al.(2021)Meister, Lazov, Augenstein, and
  Cotterell}]{meister2021sparse}
Clara Meister, Stefan Lazov, Isabelle Augenstein, and Ryan Cotterell. 2021.
\newblock Is sparse attention more interpretable?
\newblock In \emph{ACL}.

\bibitem[{Michel et~al.(2019)Michel, Levy, and Neubig}]{michel}
Paul Michel, Omer Levy, and Graham Neubig. 2019.
\newblock \href {http://arxiv.org/abs/1905.10650} {Are sixteen heads really
  better than one?}
\newblock \emph{CoRR}, abs/1905.10650.

\bibitem[{Nivre et~al.(2020)Nivre, de~Marneffe, Ginter, Hajic, Manning,
  Pyysalo, Schuster, Tyers, and Zeman}]{dependencies}
Joakim Nivre, Marie{-}Catherine de~Marneffe, Filip Ginter, Jan Hajic,
  Christopher~D. Manning, Sampo Pyysalo, Sebastian Schuster, Francis~M. Tyers,
  and Daniel Zeman. 2020.
\newblock \href {http://arxiv.org/abs/2004.10643} {Universal dependencies v2:
  An evergrowing multilingual treebank collection}.
\newblock \emph{CoRR}, abs/2004.10643.

\bibitem[{Parikh et~al.(2016)Parikh, T{\"a}ckstr{\"o}m, Das, and
  Uszkoreit}]{attention_ex2}
Ankur Parikh, Oscar T{\"a}ckstr{\"o}m, Dipanjan Das, and Jakob Uszkoreit. 2016.
\newblock \href {https://doi.org/10.18653/v1/D16-1244} {A decomposable
  attention model for natural language inference}.
\newblock In \emph{Proceedings of the 2016 Conference on Empirical Methods in
  Natural Language Processing}, pages 2249--2255, Austin, Texas. Association
  for Computational Linguistics.

\bibitem[{Paulus et~al.(2017)Paulus, Xiong, and Socher}]{attention_ex3}
Romain Paulus, Caiming Xiong, and Richard Socher. 2017.
\newblock \href {http://arxiv.org/abs/1705.04304} {A deep reinforced model for
  abstractive summarization}.
\newblock \emph{CoRR}, abs/1705.04304.

\bibitem[{Pennington et~al.(2014)Pennington, Socher, and
  Manning}]{pennington-etal-2014-glove}
Jeffrey Pennington, Richard Socher, and Christopher Manning. 2014.
\newblock \href {https://doi.org/10.3115/v1/D14-1162} {{G}lo{V}e: Global
  vectors for word representation}.
\newblock In \emph{Proceedings of the 2014 Conference on Empirical Methods in
  Natural Language Processing ({EMNLP})}, pages 1532--1543, Doha, Qatar.
  Association for Computational Linguistics.

\bibitem[{Qiu et~al.(2019)Qiu, Ma, Levy, Yih, Wang, and
  Tang}]{qiu2019blockwise}
Jiezhong Qiu, Hao Ma, Omer Levy, Scott Wen-tau Yih, Sinong Wang, and Jie Tang.
  2019.
\newblock Blockwise self-attention for long document understanding.
\newblock \emph{arXiv preprint arXiv:1911.02972}.

\bibitem[{Ren et~al.(2021)Ren, Dai, Dai, Yang, Leskovec, Schuurmans, and
  Dai}]{ren2021combiner}
Hongyu Ren, Hanjun Dai, Zihang Dai, Mengjiao Yang, Jure Leskovec, Dale
  Schuurmans, and Bo~Dai. 2021.
\newblock Combiner: Full attention transformer with sparse computation cost.
\newblock In \emph{NeurIPS}.

\bibitem[{Richardson and Urbanke(2008)}]{codingtheorybook}
Tom Richardson and Ruediger Urbanke. 2008.
\newblock \emph{Modern Coding Theory}.
\newblock Cambridge University Press.

\bibitem[{Roy et~al.(2020)Roy, Saffar, Vaswani, and Grangier}]{Roy}
Aurko Roy, Mohammad Saffar, Ashish Vaswani, and David Grangier. 2020.
\newblock \href {http://arxiv.org/abs/2003.05997} {Efficient content-based
  sparse attention with routing transformers}.
\newblock \emph{CoRR}, abs/2003.05997.

\bibitem[{Roy et~al.(2021)Roy, Saffar, Vaswani, and
  Grangier}]{roy2021efficient}
Aurko Roy, Mohammad Saffar, Ashish Vaswani, and David Grangier. 2021.
\newblock Efficient content-based sparse attention with routing transformers.
\newblock \emph{Transactions of the Association for Computational Linguistics},
  9:53--68.

\bibitem[{Sukhbaatar et~al.(2019)Sukhbaatar, Grave, Bojanowski, and
  Joulin}]{sukhbaatar2019adaptive}
Sainbayar Sukhbaatar, Edouard Grave, Piotr Bojanowski, and Armand Joulin. 2019.
\newblock Adaptive attention span in transformers.
\newblock \emph{arXiv preprint arXiv:1905.07799}.

\bibitem[{Tenney et~al.(2019)Tenney, Das, and Pavlick}]{tenney}
Ian Tenney, Dipanjan Das, and Ellie Pavlick. 2019.
\newblock \href {http://arxiv.org/abs/1905.05950} {{BERT} rediscovers the
  classical {NLP} pipeline}.
\newblock \emph{CoRR}, abs/1905.05950.

\bibitem[{Vaswani et~al.(2017)Vaswani, Shazeer, Parmar, Uszkoreit, Jones,
  Gomez, Kaiser, and Polosukhin}]{vaswani2017attention}
Ashish Vaswani, Noam Shazeer, Niki Parmar, Jakob Uszkoreit, Llion Jones,
  Aidan~N Gomez, {\L}ukasz Kaiser, and Illia Polosukhin. 2017.
\newblock Attention is all you need.
\newblock In \emph{NIPS}, pages 5998--6008.

\bibitem[{Voita et~al.(2019)Voita, Talbot, Moiseev, Sennrich, and
  Titov}]{voita-etal-2019-analyzing}
Elena Voita, David Talbot, Fedor Moiseev, Rico Sennrich, and Ivan Titov. 2019.
\newblock \href {https://doi.org/10.18653/v1/P19-1580} {Analyzing multi-head
  self-attention: Specialized heads do the heavy lifting, the rest can be
  pruned}.
\newblock In \emph{Proceedings of the 57th Annual Meeting of the Association
  for Computational Linguistics}, pages 5797--5808, Florence, Italy.
  Association for Computational Linguistics.

\bibitem[{Wang et~al.(2018)Wang, Singh, Michael, Hill, Levy, and Bowman}]{glue}
Alex Wang, Amanpreet Singh, Julian Michael, Felix Hill, Omer Levy, and
  Samuel~R. Bowman. 2018.
\newblock \href {http://arxiv.org/abs/1804.07461} {{GLUE:} {A} multi-task
  benchmark and analysis platform for natural language understanding}.
\newblock \emph{CoRR}, abs/1804.07461.

\bibitem[{Zaheer et~al.(2020)Zaheer, Guruganesh, Dubey, Ainslie, Alberti,
  Onta{\~{n}}{\'{o}}n, Pham, Ravula, Wang, Yang, and Ahmed}]{zaheer}
Manzil Zaheer, Guru Guruganesh, Avinava Dubey, Joshua Ainslie, Chris Alberti,
  Santiago Onta{\~{n}}{\'{o}}n, Philip Pham, Anirudh Ravula, Qifan Wang,
  Li~Yang, and Amr Ahmed. 2020.
\newblock \href {http://arxiv.org/abs/2007.14062} {Big bird: Transformers for
  longer sequences}.
\newblock \emph{CoRR}, abs/2007.14062.

\bibitem[{Zanzotto et~al.(2020)Zanzotto, Santilli, Ranaldi, Onorati, Tommasino,
  and Fallucchi}]{kermit}
Fabio~Massimo Zanzotto, Andrea Santilli, Leonardo Ranaldi, Dario Onorati,
  Pierfrancesco Tommasino, and Francesca Fallucchi. 2020.
\newblock \href {https://doi.org/10.18653/v1/2020.emnlp-main.18} {{KERMIT}:
  Complementing transformer architectures with encoders of explicit syntactic
  interpretations}.
\newblock In \emph{Proceedings of the 2020 Conference on Empirical Methods in
  Natural Language Processing (EMNLP)}, pages 256--267, Online. Association for
  Computational Linguistics.

\bibitem[{Zhang et~al.(2016)Zhang, Lu, and Lapata}]{zhang-etal-2016-top}
Xingxing Zhang, Liang Lu, and Mirella Lapata. 2016.
\newblock \href {https://doi.org/10.18653/v1/N16-1035} {Top-down tree long
  short-term memory networks}.
\newblock In \emph{Proceedings of the 2016 Conference of the North {A}merican
  Chapter of the Association for Computational Linguistics: Human Language
  Technologies}, pages 310--320, San Diego, California. Association for
  Computational Linguistics.

\end{thebibliography}
\bibliographystyle{acl_natbib}

\appendix

\section{Appendix}
\label{sec:appendix}

\subsection{Finetuning BERT}
\label{finetune_hparam}
For all datasets we use the BERT Base model: 12 layers, 12 heads and a hidden dimension of 768, containing $\approx$ 109M parameters. We load the publicly available pretrained uncased checkpoint before finetuning on our data. We use the ADAM optimizer for 10 epochs with a learning rate of 5e-5 and a warmup ratio of 0.1, following the standard practice of finetuning BERT~\cite{bert}. We set the train batch size to 32.

\subsection{Pre-training with learned sparsity}
\begin{figure*}[h!]
\centering
\begin{subfigure}[b]{0.45\textwidth}
  \centering
   \raisebox{3mm}{\includegraphics[width=0.9\linewidth]{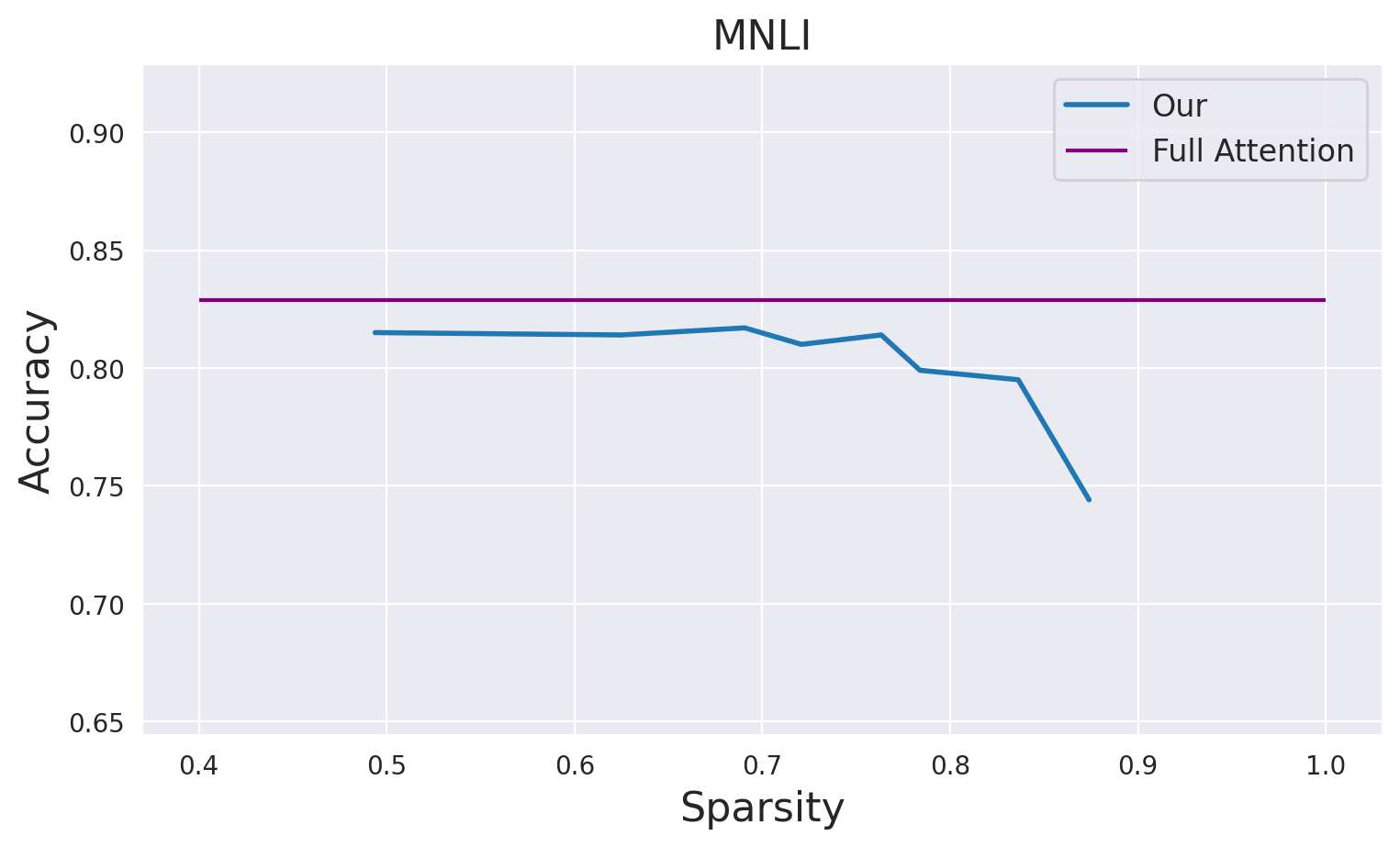}}
   \caption{Comparison between our and full model accuracy.}
 \end{subfigure}%
 \begin{subfigure}[b]{0.45\textwidth}
  \centering
   \includegraphics[width=0.9\linewidth]{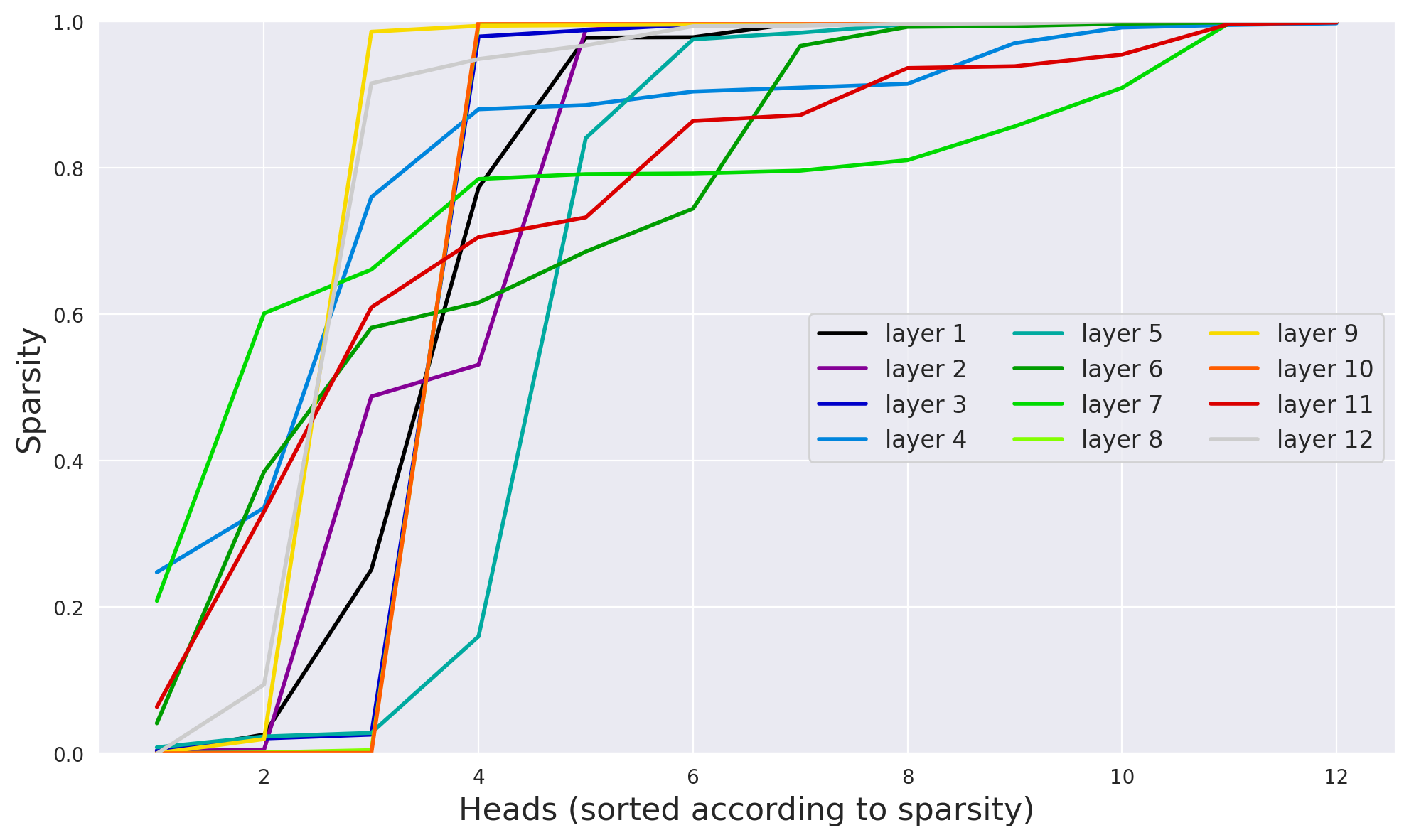}
   \caption{Sparsity of heads in each layer. (overall sparsity=0.763)}
 \end{subfigure}
 \caption{Results for the Learned Sparsity framework for pretraining + finetuning on the MNLI dataset. (a) Accuracy is close to full attention performance up to almost 70\% sparsity, which demonstrates the effectiveness of our method. (b) Distribution of the sparsities for all the attention heads showing the sparsity patterns emerging in the heads and can help identify the underlying learning dynamics.}
 \label{fig:pretraining}
 \end{figure*}

We pre-train sparse BERT-Base models from scratch, optimizing for a loss similar to \eqref{eq:degopt} with the addition of an MLM loss term. We follow the standard process of BERT pre-training \cite{bert}.  To make the training more stable, we use a slightly different parameterization of $\delta_h^{\ell}$ than the sigmoid() of free variable used in Sec \ref{lsdd}. This is followed by finetuning on the MNLI dataset. As shown in Fig. \ref{fig:pretraining}(a), the models pre-trained with learned sparsity achieve accuracies  close to full attention performance even at >70\% sparsity, one model achieving an accuracy of 0.814 (compared to 0.829 of the full model) at a sparsity of 0.763. This shows the effectiveness of our approach.
In Fig. \ref{fig:pretraining}(b), we show the sparsities of the neighbour patterns learned by each head in each layer. It is interesting to see the diversity of these patterns across the layers.  

\subsection{Computation Budget}
For our experiments we used a total of $\approx$ 800 hours on TPUv2 (8-core) accelerators.

\subsection{Limitations}
We conduct experiments on only 3 finetuning tasks, all in English. More experiments need to be done to confirm our results on other tasks or datasets with a greater diversity of languages. Due to the large number of experiments, we limited our choice of hyperparameters to the standard ones used for finetuning BERT in the literature. Further, our study is restricted to the standard transformer model used in BERT and further studies are required to generalize our results to the numerous other transformer variants used in the practice.
\end{document}